%% file: hyped.tex
\documentclass[10pt,twocolumn,letterpaper]{article}
\pdfoutput=1

\usepackage{iccv}
\usepackage{times}
\usepackage{epsfig}
\usepackage{graphicx}
\usepackage{amsmath}
\usepackage{amssymb}

\usepackage[breaklinks=true,bookmarks=false]{hyperref}

\iccvfinalcopy
\def\iccvPaperID{512}

\ificcvfinal\fi
\begin{document}

\title{Learned Spectral Super-Resolution}

\author{Silvano Galliani$^1$ \hspace{2em} Charis Lanaras$^1$ \hspace{2em} Dimitrios Marmanis$^2$ \\ \hspace{2em} Emmanuel Baltsavias$^1$ \hspace{1em} Konrad Schindler$^1$
  \vspace{1em}
  \\
$^1$Photogrammetry and Remote Sensing, ETH Zurich, Switzerland\\
$^2$DLR-IMF Department, German Aerospace Center, Oberpfaffenhofen, Germany
}

\maketitle
\ificcvfinal\thispagestyle{empty}\fi

\input{abstract}

\input{introduction}

\input{relatedworks}

\input{method}

\input{results}
\input{conclusion}

{\small
\bibliographystyle{ieee}
\bibliography{hyped-cnn}
}

\end{document}

%% file: abstract.tex
\begin{abstract}
  We describe a novel method for blind, single-image
  \mbox{\textbf{spectral}} super-resolution.  While conventional
  super-resolution aims to increase the spatial resolution of an input
  image, our goal is to spectrally enhance the input, \ie, generate an
  image with the same spatial resolution, but a greatly increased
  number of narrow (hyper-spectral) wavelength bands. Just like the
  spatial statistics of natural images has rich structure, which one
  can exploit as prior to predict high-frequency content from a low
  resolution image, the same is also true in the spectral domain: the
  materials and lighting conditions of the observed world induce
  structure in the spectrum of wavelengths observed at a given
  pixel. Surprisingly, very little work exists that attempts to use
  this diagnosis and achieve blind spectral super-resolution from
  single images. We start from the conjecture that, just like in the
  spatial domain, we can learn the statistics of natural image
  spectra, and with its help generate finely resolved hyper-spectral
  images from RGB input. Technically, we follow the current best
  practice and implement a convolutional neural network (CNN), which
  is trained to carry out the end-to-end mapping from an entire RGB
  image to the corresponding hyperspectral image of equal size. We
  demonstrate spectral super-resolution both for conventional RGB
  images and for multi-spectral satellite data, outperforming the
  state-of-the-art.
\end{abstract}

%% file: introduction.tex
\section{Introduction}
Single-image super-resolution is a challenging computer vision  problem
with many interesting applications, \eg~in the fields of astronomy, medical
imaging and law enforcement.
The goal is to infer, from a single low-resolution image, the missing
high frequency content that would be visible in a corresponding high
resolution image.
The problem itself is inherently ill-posed, extremely so for large
upscaling factors. Still, several successful schemes have been
designed~\cite{dong2011image}.
The key is to exploit the high degree of structure in the visual world
and design or learn a prior that constrains the solution accordingly.

\begin{figure}[t]
\begin{center}
\includegraphics[width=1\columnwidth]{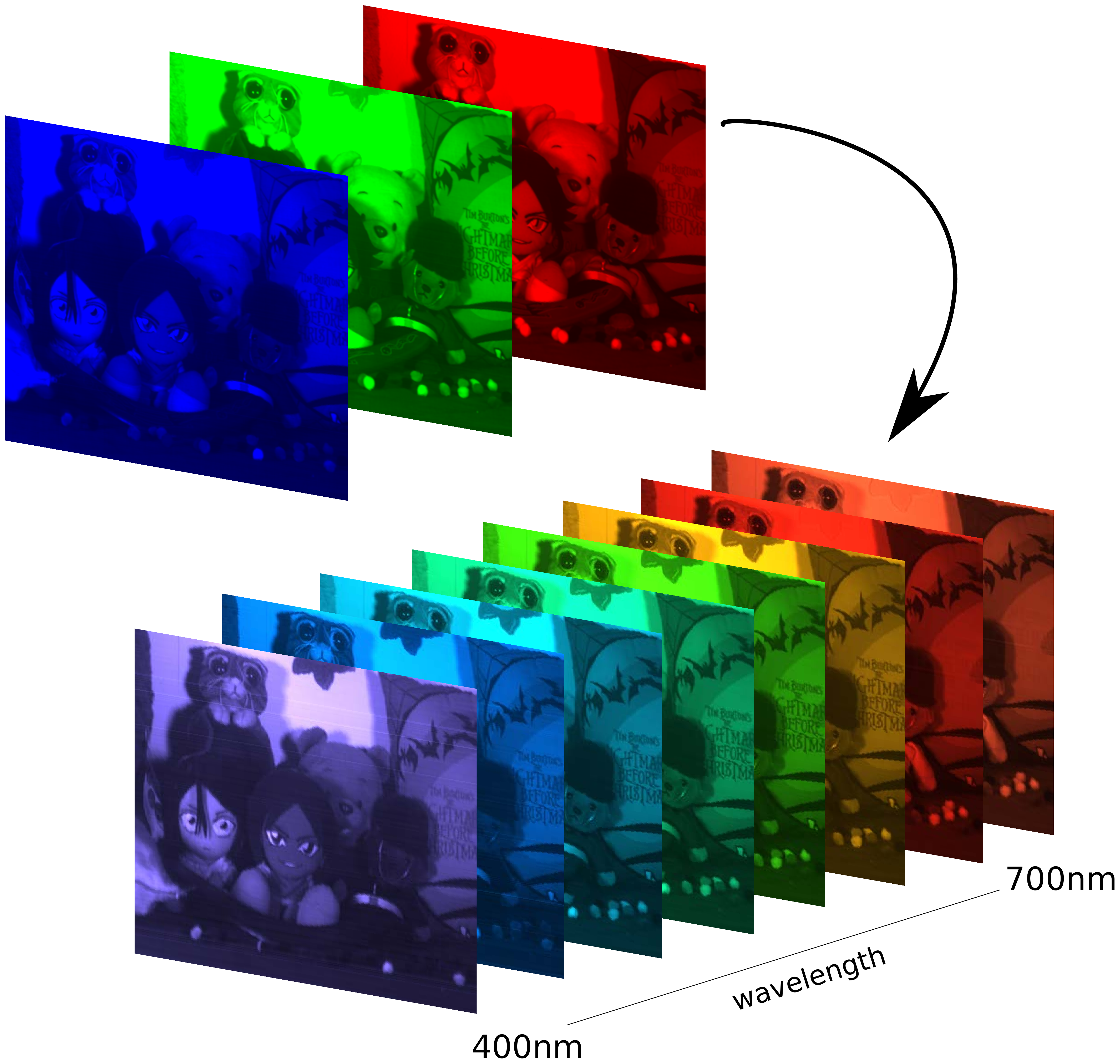}
\caption{Spectral super-resolution: our method is able to predict fine-grained
hyperspectral images, using only
a single RGB image as input (number of output channels reduced for visualisation, actual output has 31 bands of width 10 nm).}
\label{fig:teaser}
\end{center}
\end{figure}

Indeed there is a large body of literature on single-image
super-resolution, which is however largely limited to the spatial
domain.
Very few authors address the complementary problem, to increase the
\emph{spectral} resolution of the input image beyond the coarse RGB
channels.
The topic of this paper is single-image \emph{spectral
  super-resolution}. We pose the obvious question whether we can also
learn the spectral structure of the visual world, and use it as a
prior to predict hyper-spectral images with finer spectral resolution
from a standard RGB image.\footnote{Or from some other image with similarly broad channels, \eg, a color infrared image.} Note the trade-off between spatial and spectral information, even at
the sensor level: to obtain a reasonable signal-to-noise ratio,
cameras can have small pixels and integrate over large spectral bands;
or they can have fine spectral resolution, but integrate over large
pixels.

Depending on the available images and the application, it may be
useful to increase the resolution in space or to obtain a finer
quantisation of the visible spectrum.
While in the spatial domain the restoration of missing high-frequency
information reveals smaller objects and more accurate boundaries,
high-frequency spectral information makes it easier to separate the
\textit{spectral signatures} of different objects and materials that
have similar RGB color.
The extra information included in the recovered hyper-spectral (HS)
image bands enables applications like tracking~\cite{van2010tracking},
segmentation~\cite{tarabalka2010segmentation}, face
recognition~\cite{pan2003face}, document
analysis~\cite{kim2011visual,padoan2008quantitative}, analysis of
paintings~\cite{elias2008multispectral}, food
inspection~\cite{wu2013advanced} and image
classification~\cite{camps2013advances}.

A related, but simpler problem has been studied by several authors,
namely \textit{hyper-spectral super-resolution}~\cite{akhtar2016hierarchical,
kawakami2011high,lanaras2015hyperspectral, simoes2015convex}.
There, one assumes that both a HS image of low spatial resolution and
an RGB image with finer resolution are available, and the two are
fused to get the best of both worlds.
The desired output is thus the same as in our problem --- but requires
an additional input.
Our work can be seen as an attempt to do away with the spatially
coarse hyper-spectral image and learn a generic prior for
hyper-spectral signatures.

The problem is heavily under-constrained: for typical terrestrial
applications, the goal is to generate, for each pixel, $\approx$30
spectral bands from the 3 input channels.
The difference is even more extreme in aerial and satellite remote
sensing, where the low-resolution image has at most 10 channels
covering the visible and infrared range, whereas hyper-spectral images
routinely have $>$200 bands over the same range.
Still, there is evidence that blind spectral super-resolution is
possible. For practical processing, hyper-spectral signatures are
sometimes projected to a lower-dimensional subspace
\cite{bioucas2013hyperspectral}, indicating that there is a
significant amount of correlation between their bands.
Moreover, most scenes consist of a limited number of materials,
distributed in characteristic patterns. Thus, there is hope that one
can learn them from a suitable training set.
Here, we do exactly that: we train a convolutional neural network
(CNN) to predict the missing high-frequency detail of the colour
spectrum observed at each pixel.

There are two main differences to spatial super-resolution, which has
also been tackled with CNNs.
First, spatial super-resolution has the convenient property that
training data can be obtained by downsampling existing images of the
desired resolution, so training data is available for free in
virtually unlimited quantities.
This is not the case for our problem, because hyper-spectral cameras
are not a ubiquitous consumer product, and training data is comparatively rare.
We nevertheless manage to obtain enough training data even if we are constrained
to a more limited amount of image. In cases where the overall number of images
is small we regularize the solution with an Euclidean penalized and additionally
augment the training data by flipping and rotating input images.
Second, and more importantly, the point spread functions of different cameras are rather
similar and in general steep, whereas the spectral response (the
``spectral blur kernel'') of the color channels can vary
significantly from sensor to sensor.
The latter means that an individual super-resolution has to be learned
for each camera type.

%% file: relatedworks.tex
\section{Related Work}

\emph{Single-image super-resolution} usually corresponds to spatially
upsampling a single low-resolution RGB image to higher spatial resolution.
This has been a popular topic for several years, and quite some
literature exists. Early method attempted to devise clever upsampling
functions, sometimes by manually analyzing the image statistics, while
recently the trend has been to learn dictionaries of image patches,
often in combination with a sparsity prior~\cite{timofte2013anchored,yang2010image, zeyde2010single}.
Lately, CNNs have boosted the performance of super-resolution, showing
significant improvements~\cite{dong2014learning,kim2016accurate,kim2016deeply}.
They are also able to perform the task in real time~\cite{shi2016real}.
Interestingly, the RMSE does not seem to be the best loss function to
obtain visually convincing upsampling results. Other loss functions
aiming for ``photo-realism'' better match human perception, although
the actual intensity differences are higher~\cite{ledig2016photo}.

On the other hand \emph{hyperspectral super-resolution} uses as input
a low resolution hyperspectral \emph{and} an RGB image to create a
high resolution hyperspectral output.
There are two main schools.
Some methods require only known spectral response of the RGB camera,
but can correct for spatial mis-alignment
known~\cite{akhtar2014sparse,akhtar2016hierarchical,kawakami2011high}.
Others assume that also the registration between the two input images
is perfectly known~\cite{lanaras2015hyperspectral,simoes2015convex, wei2015fast,
wycoff2013non, yokoya2012coupled}.

Our work also has some relation to the problem of image colorization,
where a grayscale image is spectrally upsampled to RGB, \ie, from one
channel to three. There CNNs have also shown promising results
\cite{larsson2016learning, zhang2016colorful} by converting the input to a
\textit{Lab} colorspace and predicting the \textit{ab} channels.

Acquiring a hyperspectral image by using only an RGB camera has been
attempted with the help of active lighting~\cite{chi2010multi}. This
can be achieved by using spectral filters in front of the
illumination, with the main disadvantage that the method can only be
used in the laboratory.
A similar idea is to use tunable narrow-band filters and take multiple
images, such that narrow spectral bands are recorded sequentially
\cite{gat2000imaging}.
Taking a step further from tuning the hardware, Wug~\etal{}
\cite{wug2016yourself} proposed the use of multiple RGB images from
different cameras, which are combined to obtain a single
hyper-spectral image --- effectively turning the differences between
the camera's spectral responses into an advantage.
All these solutions require dedicated hardware as well as a static
scene.

On the contrary, attempts to reconstruct hyper-spectral information
from a single RGB image are rare.
Nguyen \etal~\cite{nguyen2014training} use a radial basis function
network to model the mapping from RGB values to scene
reflectance. They assume the camera's spectral response function is
perfectly known (and, as a by-product, also estimate the spectral
illumination).
More recently, Arad~\etal~\cite{arad2016sparse} proposed to learn
sparse dictionary with K-SVD as hyper-spectral upsampling prior.
Assuming the spectral response of the RGB camera is known, they then
use Orthogonal Matching Pursuit (OMP) to reconstruct the hyper-spectral signal using from the RGB
intensities.
Closely related methods exist for spatial super-resolution
\cite{zeyde2010single} as well as hyper-spectral super-resolution
\cite{akhtar2014sparse}. The two methods are closely related the only main
technical difference is on which images are used to learn the dictionary.
Zeyde~\etal~\cite{zeyde2010single} employ low resolution hyperspectral image as
prior while Akhtar~\etal~\cite{akhtar2014sparse} compute their prior on a
similar image contained inside their dataset.

To summarize, several constrained versions of the spectral
super-resolution problem have been investigated. But we believe that
our work is the first generic framework that requires only a single
RGB image, no knowledge of the spectral response functions, can be
used indoors and outdoors, and needs neither a static scene nor
special filter hardware.

%% file: method.tex
\begin{figure}[t]
\begin{center}
\includegraphics[width=1\columnwidth]{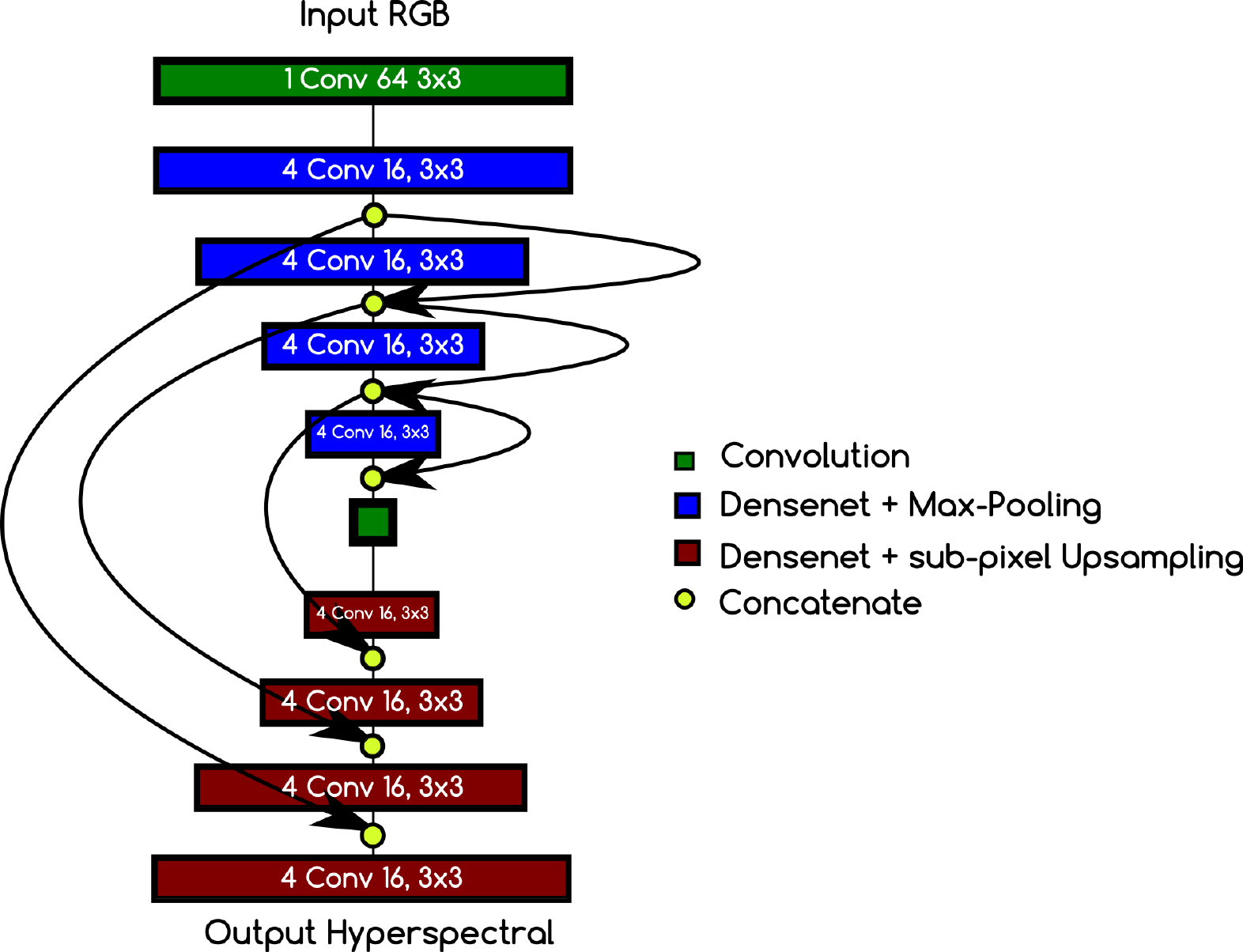}
\caption{Diagram of our network for spectral super resolution. Skip
  connection propagate the information copying and concatenating the
  output from earlier layers. The multi scale structure allows to
  explore the whole spatial extent of the input image. Note that,
  except for the first convolutions, the other blocks are made of a
dense block as in~\cite{jegou2016one} }
\label{fig:tiramisu}
\end{center}
\end{figure}

\begin{figure}[t]
\begin{center}
\includegraphics[width=0.4\columnwidth]{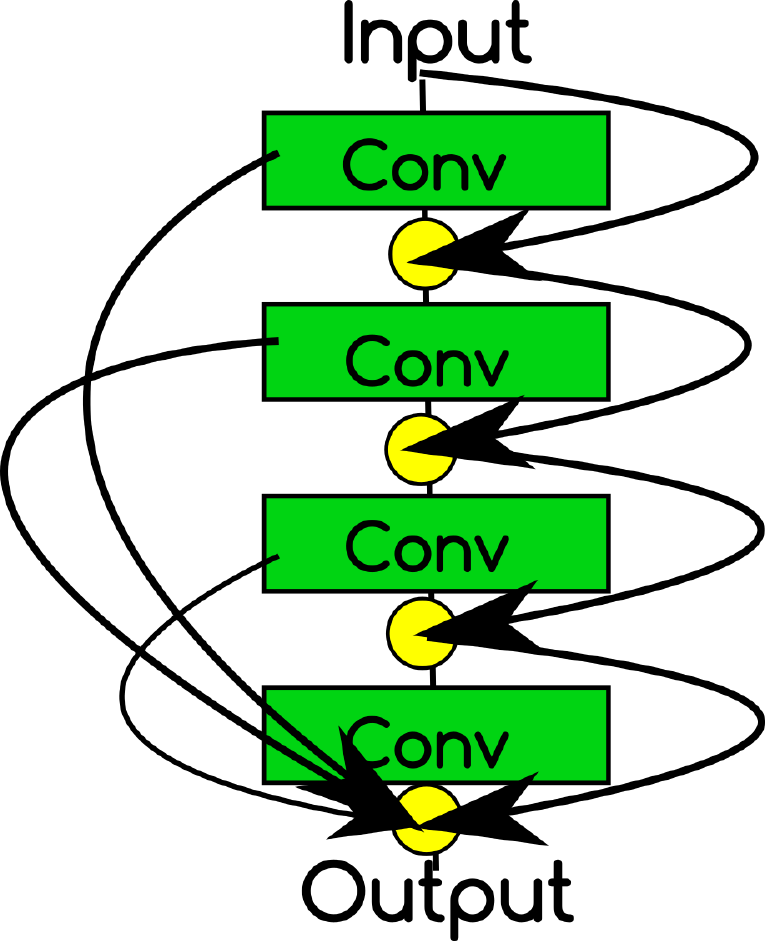}
\caption{Depiction of a single Densenet block}
\label{fig:densenet}
\end{center}
\end{figure}
\section{Method}

In our work we follow the current rend in computer vision research and
learn the desired super-resolution mapping end-to-end with a
(convolutonal) neural network.
In the following we present the network architecture and give
implementation details.

\begin{figure*}[t]
\begin{center}
\includegraphics[width=0.8\textwidth]{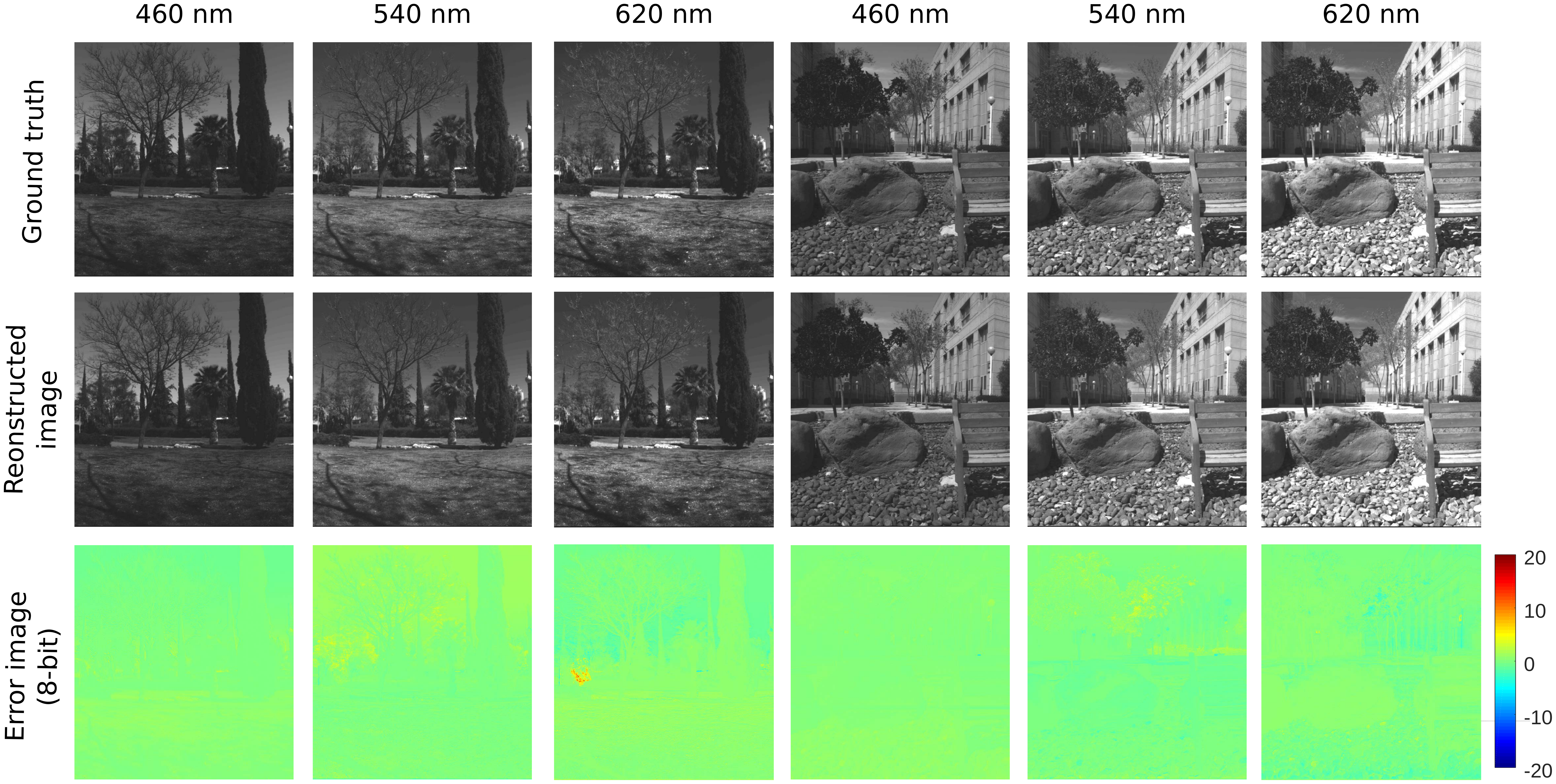}
\caption{Comparison of reconstructed radiance to the ground truth values. Two images from ICVL dataset with the same range as presented by Arad and Ben-Shahar \etal \cite{arad2016sparse}}
\label{fig:errrorsICVL}
\end{center}
\end{figure*}

Selecting a network architecture for deep learning is not
straightforward for a novel application, where no prior studies point
to suitable designs.
It is however clear that, for our purposes, the output should have the
same image size as the input (but more channels). We thus build on
recent work in semantic segmentation.
Our proposed network is a variant of the semantic segmentation
architecture \textit{Tiramisu} of J{\'e}gou~\etal\cite{jegou2016one},
which in turn is based on the
\textit{Densenet}~\cite{huang2016densely} architecture for
classification.
As a first measure, we replace the loss function and use an Euclidean
loss, since we face a regression-type problem: instead of class labels
(respectively, class scores) our network shall predict the
continuously varying intensities for all spectral bands.
Additionally, since we are interested in the high fidelity
representation of each pixel we replace the original deconvolution
layer with subpixel upsampling as proposed by the super-resolution
work of Shi~\etal~\cite{shi2016real}.
%

%
The \textit{Tiramisu} network has further interesting properties for
our task.  Skip connections, within and across Densenet blocks (see
\ref{fig:tiramisu}, \ref{fig:densenet}) perform concatenation
instead of summation of layers, as opposed to ResNet~\cite{He2015}.
They greatly speed up the learning and alleviate the vanishing
gradient problem.  More importantly, its architecture is based on a
multiscale paradigm which allows the network to learn the overall
image structure, while keeping the image resolution constant.
In the \textit{Tiramisu} structure, each downscaling step is done with
a convolutional layer of size $1$ and $\max$-pooling, while for each
resolution level a single \textit{Densenet} is used, with varying
number of convolutional layers.

Our network architecture, with a total of 56 layers, is depicted in
Fig.~\ref{fig:tiramisu} where, if otherwise specified, each convolution
has size $3\times3$.

The image gets down-scaled 5 times by a factor of 2, with a $1\times1$
convolution followed by $\max$-pooling.  In it's own terminology, each
\emph{Densenet} block has a growth rate of 4 with 16 layers, which
means 4 convolutional layers per block, each with 16 filters,
see Fig.~\ref{fig:densenet}.  For a more details about the
\emph{Densenet/Tiramisu} architecture, please refer to the original
papers~\cite{huang2016densely, jegou2016one}.

For each image in the training dataset we randomly sample a set of
patches of size $64\times64$ and directly feed them to the neural
network.  At test time, where the goal is to reconstruct the complete
image, we tile the input into $64\times64$ tiles, with 8 pixels
overlap to avoid boundary artifacts.


\subsection{Relation to spectral unmixing}
Often, hyper-spectral images are interpreted in terms of
``endmembers'' and ``abundances'': the endmembers can be imagined as
the pure spectra of the observed materials and form a natural
basis. Observed pixel spectra are additive combinations of endmembers,
with the abundances (proportions of different materials) as
coefficients.

Dong~\etal~\cite{dong2014learning} showed how a shallow CNN for
super-resolution can be interpreted in terms of basis learning and
reconstruction.
In much the same manner, our CNN can be seen as an implicit,
non-linear extension of the unmixing model, where the knowledge about
the endmembers at both low and high spectral resolution is embedded in
the convolution weights.
The forward pass through the network can be thought of as first
extracting the abundances from the input and then multiplying them
with the learned endmembers to obtain the hyperspectral output image.


\subsection{Implementation details}
The network, implemented with Keras~\cite{chollet2015keras}, is
trained from scratch, using the Adam optimizer~\cite{kingma2014adam}
with Nesterov moment~\cite{sutskever2013importance,
  dozat2015incorporating}. We iterate it for $100$ epochs with
learning rate $0.002$, then for another $200$ epoch with learning rate
$0.0002$, the rest of the parameters follows those provided in the
paper. We initialize our model with
\emph{HeUniform}~\cite{he2015delving}, and apply $50\%$ dropout in the
convolutional layers to avoid overfitting.  Moreover, we found it
crucial to carefully tune the Euclidean regularization, probably due
to the general lack of copious amount of images on the training
set. We fix it to $10^{-6}$, higher values lead to overly smooth, less
accurate solutions.

%% file: results.tex
\begin{figure*}[t]
\begin{center}
\includegraphics[trim={0 0 0 0},clip,width=0.7\linewidth]{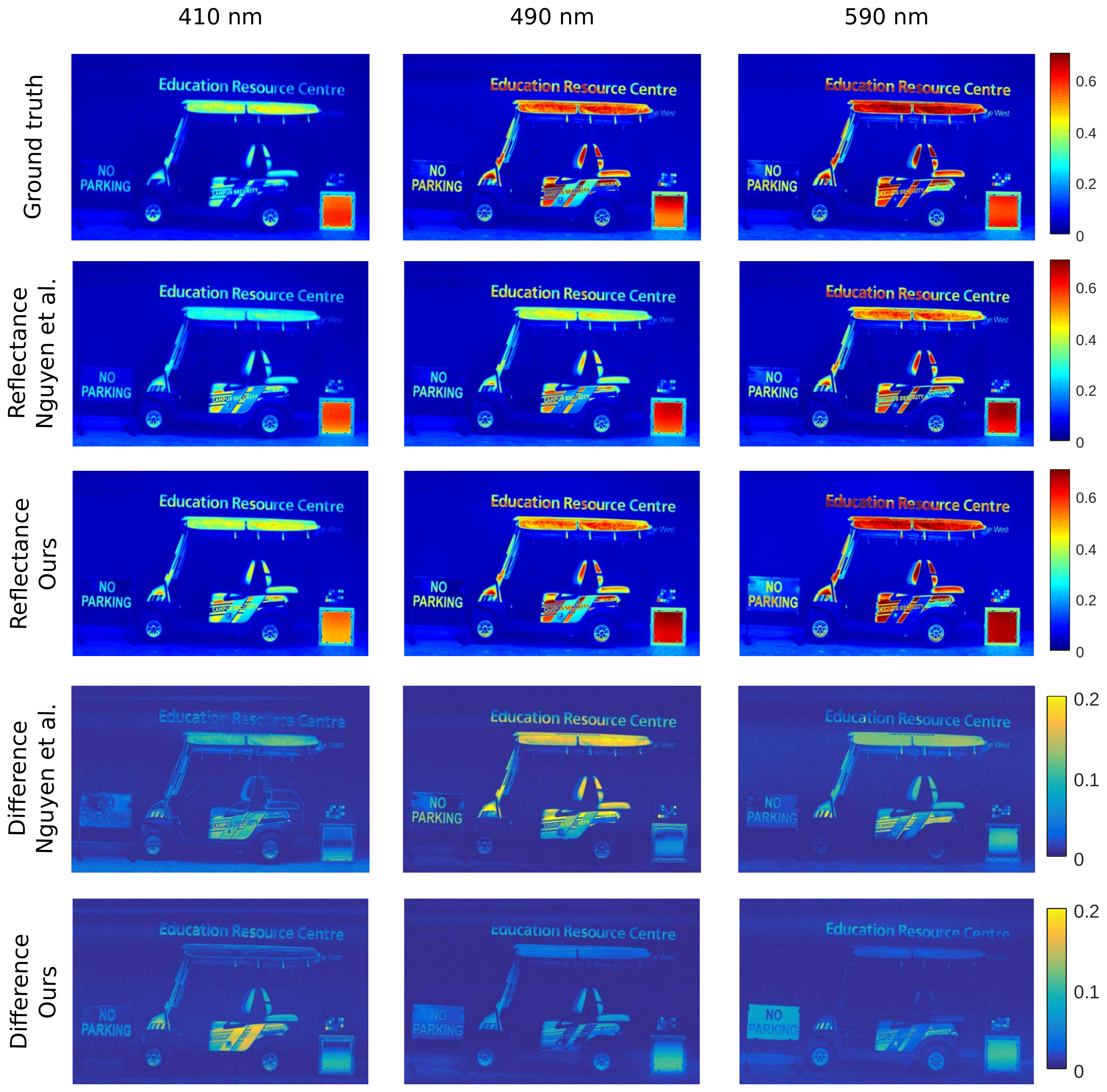}
\caption{Qualitative comparison~\wrt~\cite{nguyen2014training} over three non consecutive spectral bands.}
\label{fig:diffNUS}
\end{center}
\end{figure*}

\section{Results}
We evaluate our result on four different datasets. Where possible, we
compare it with the other two methods~\cite{arad2016sparse,
  nguyen2014training} that are also able to estimate an
hyperspectral image from an single RGB.
Note, both baselines need the spectral response of the RGB camera to
be known. Hence, we feed it to them as additional input, contrary to
our method. Despite the disadvantage, our CNN super-resolution is more
accurate, see below.
\paragraph{Error computation}
We evaluate \wrt three different error metric over 8bit images (as far as they are
available): root mean square error (RMSE), relative root mean square
error (RMSERel), and the spectral angle mapper (SAM)
\cite{yuhas1992discrimination}, \ie, the average angular deviation
between the estimated pixel spectra, measured in degrees.
We would like to highlight how we measure RMSERel and RMSE as there is not a
common agreement on its computation. RMSE is obtained by computing it on 8bit
and clipping values higher and lower than the allowed range. For RMSERel we
normalized the predicted image by the mean of the ground truth.

\subsection{Training data}

We follow the standard practice for quantitative evaluation and
synthetically generate the input data, given the difficulties of
capturing separate hyper-spectral and RGB images that are aligned and
have comparable resolution and sharpness.
\Ie, the RGB image is emulated by integrating over hyper-spectral
channels according to a predefined camera response function.
We always use the response functions provided by the authors, to
ensure the images are strictly the same and the comparisons are fair.
If a dataset already provides a train/test split, we follow
it. Otherwise, we run two-fold cross-validation: split the dataset in
two, train on the first half to predict the second half and vice
versa.

\input{tableICVL-CAVE}

\subsection{ICVL dataset}
The ICVL dataset has been released by Arad and
Ben-Shahar~\cite{arad2016sparse}, together with their method. It
contains 201 images acquired using a line scanner camera (Specim PS
Kappa DX4 hyperspectral), mounted on a rotary stage for spatial
scanning.  The dataset contains a variety of scenes captured both
indoors and outdoors, including man-made to natural objects.  Images
were originally captured with a spatial resolution of 1392$\times$1300
over 519 spectral bands (400-1,000nm ) but have been downsampled to 31
spectral channels from 400nm to 700nm at 10nm increments.
We map the hyperspectral images to RGB using CIE 1964 color
matching functions, like in the original paper.

There, a sparse dictionary is learned with K-SVD as hyper-spectral
upsampling prior.
However, they do not use a global train/test split, as we do. Rather,
they divide the dataset into subsets of images that show the same type
of scene (such as parks or indoor environments), hold out one test
image per subset, and train on the remaining ones; thus learning a
different prior for each test image that is specifically tuned to the
scene type.
%
%
We prefer to keep the prior generic and use a single, global
train/test split. We then predict their held-out images, but using the
same network for all that fall into the first, respectively second
half of our split.
Even so, our results are competitive, see Table~\ref{table:ICVL}.
We are not able to reproduce their method due to missing parameter or availability of code, instead we show the same figures presented in their paper in Fig.~\ref{fig:errrorsICVL}.

\begin{figure}[t]
\begin{center}
\includegraphics[width=1\columnwidth]{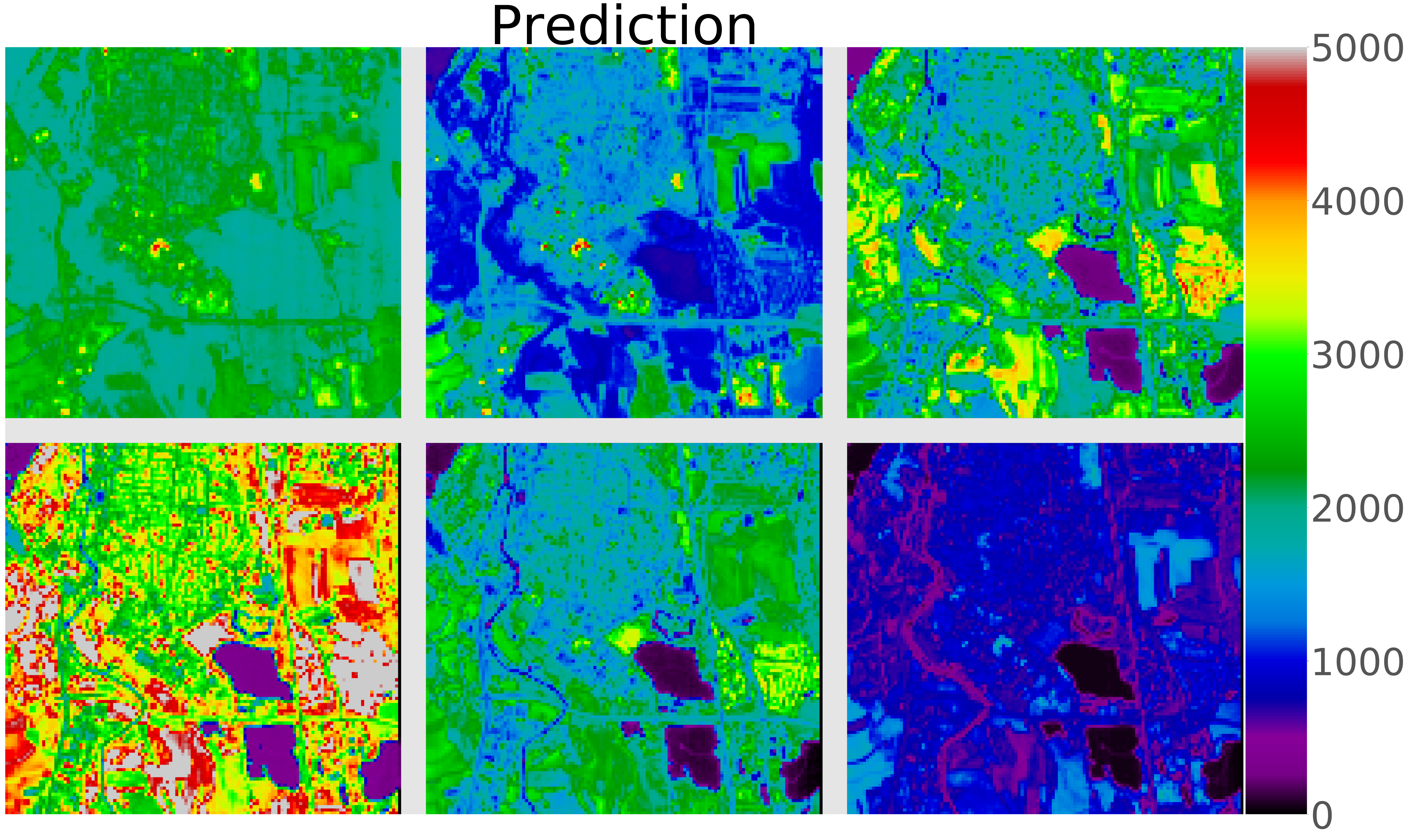}
\includegraphics[width=1\columnwidth]{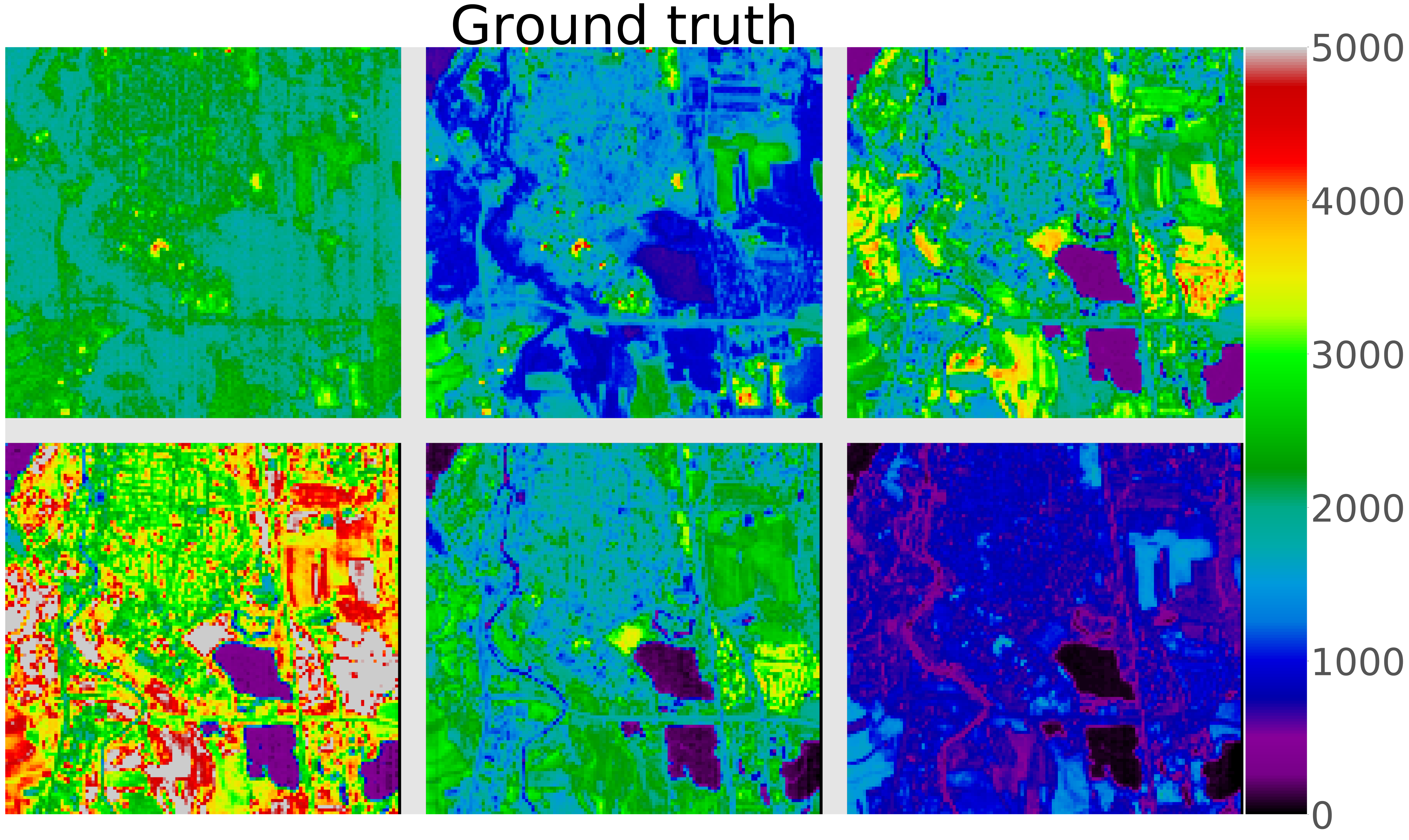}
\includegraphics[width=1\columnwidth]{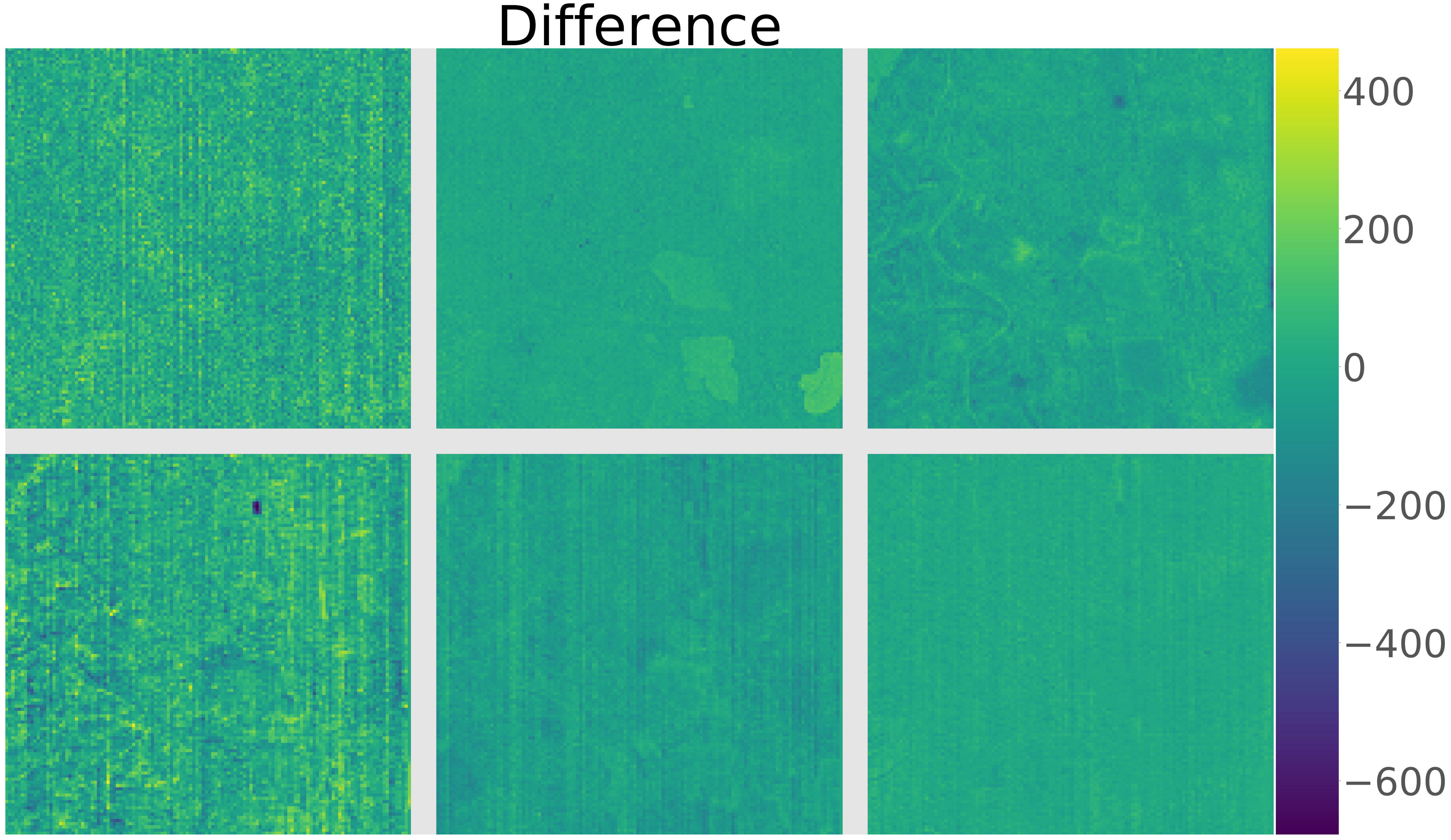}
\caption{Prediction from spectral image, ground truth and difference images for a subregion of a Hyperion satellite image. The bands shown are 0,20,40
(first row), 60,80,120 (second row), corresponding to the following
central frequencies in nm: 426, 630, 833, 1013,
1215, 1618.
Note the reconstruction quality across different bands, and that the
difference images are in fact dominated by typical sensor noise
patterns like streaking artifacts.}
\label{fig:hyperion1}
\vspace{0.5em}
\end{center}
\end{figure}

\input{tableNUSrad}

\input{tableNUSrefl}

\begin{figure*}[t]
\begin{center}
\includegraphics[width=0.9\textwidth]{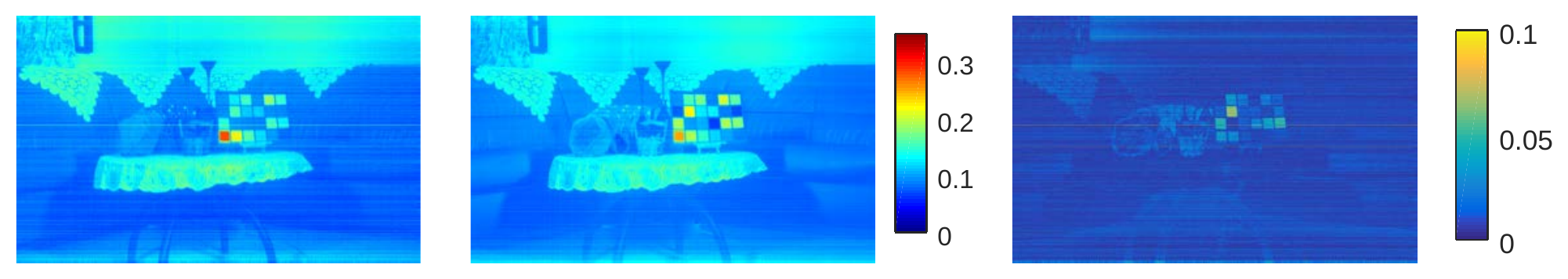}
\caption{Ground truth, prediction and difference of one image of NUS dataset. Note how the marked line artifacts in the ground truth gets removed by our method.}
\label{fig:noiseNUS}
\end{center}
\end{figure*}

\subsection{NUS dataset}
The NUS dataset~\cite{nguyen2014training} contains the spectral
irradiance and spectral illumination (400-700 nm with step of 10 nm)
for 66 outdoor and indoor scenes, captured under several different
illuminations. In that dataset the authors already prescribe a
train/test split.
What their learning method does is to estimate both the reflectance
and the illumination from an RGB image with known camera response
function.
In order to fairly evaluate their method, we run the authors' original
code to estimate the reflectance, and convert it to radiance with the
ground truth illumination.
Of the three different camera response functions evaluated in their
paper, we pick the one that gave them the best results (Canon 1D Mark III), to create the
RGB images.
Additionally we apply their ground truth illumination to our result to
also compare reflectance.
Also for this dataset our method obtain the best result in terms of
RMSE, see Tables~\ref{table:NUSradiance} and~\ref{table:NUSreflectance}.
In this case our SAM error was slightly worse, probably due to outlier on same
channels which would increase considerably the error result.

\subsection{CAVE dataset}
The CAVE dataset~\cite{yasuma2008generalized} is a popular
hyper-spectral dataset.  As opposed to all the other ones it is not
captured with a rotating line scanner. Instead, the hyper-spectral
bands are recorded sequentially with a tunable filter.
The main benefit is the elimination of possible noise when using a
\textit{pushbroom} scanner, while moving objects such as trees pose problems,
because the bands are not correctly aligned.
The dataset contains a variety of challenging objects to predict. The
heterogeneity of the captured scenes makes it harder to learn a global
prior for all scenes and challenges learning-based methods, like ours.
Nevertheless, our method is competitive \wrt the number provided by
\cite{arad2016sparse}, see Table~\ref{table:ICVL}.

\input{tableHyperion}

\subsection{Satellite Data}
We tested our method also on data captured from Hyperion
satellite~\cite{pearlman2003hyperion} a sensor on board the satellite EO-1.  The satellite carries a
hyperspectral line scanner that records 242 channels (from 0.4 to 2.5
µm) at 30 m ground resolution, out of which 198 are calibrated and can be used. Our scenes are already cloud-free, have
a size of $\approx$256$\times$7000 pixels, and show the river Rhine in
Western Europe. Note, like most satellite data the images are stored
with an intensity range of 16 bits ber channel, and have an effective
radiometric depth of$\approx$5000 different gray values. The input
image is emulated by integrating the hyper-spectral bands into the
channels of ALI, the 9-channel multispectral sensor on board the same satellite.
As test bed, we use different acquisitions dates over (roughly) the
same area.  This is of course a favourable scenario for our method:
since training and test data show the same region, the network can
learn adapt to the specific structure present in the region, and
potentially to some degree even to the scene layout.
Indeed, both the quantitative results in table~\ref{table:Hyperion} and
the visual examples in Fig~\ref{fig:hyperion1} validate the
performance of our method over multiple visible and non-visible bands.
While the training data is certainly favourable, it is not an
unrealistic assumption that legacy hyper-spectral data for a given
region is available.
We find it quite remarkable that, according to the example, we are
able to predict, with high accuracy, a finely resolved spectrum $>200$
bands from a standard, multi-spectral satellite image.

\begin{figure}[t]
\begin{center}
\includegraphics[width=1\columnwidth]{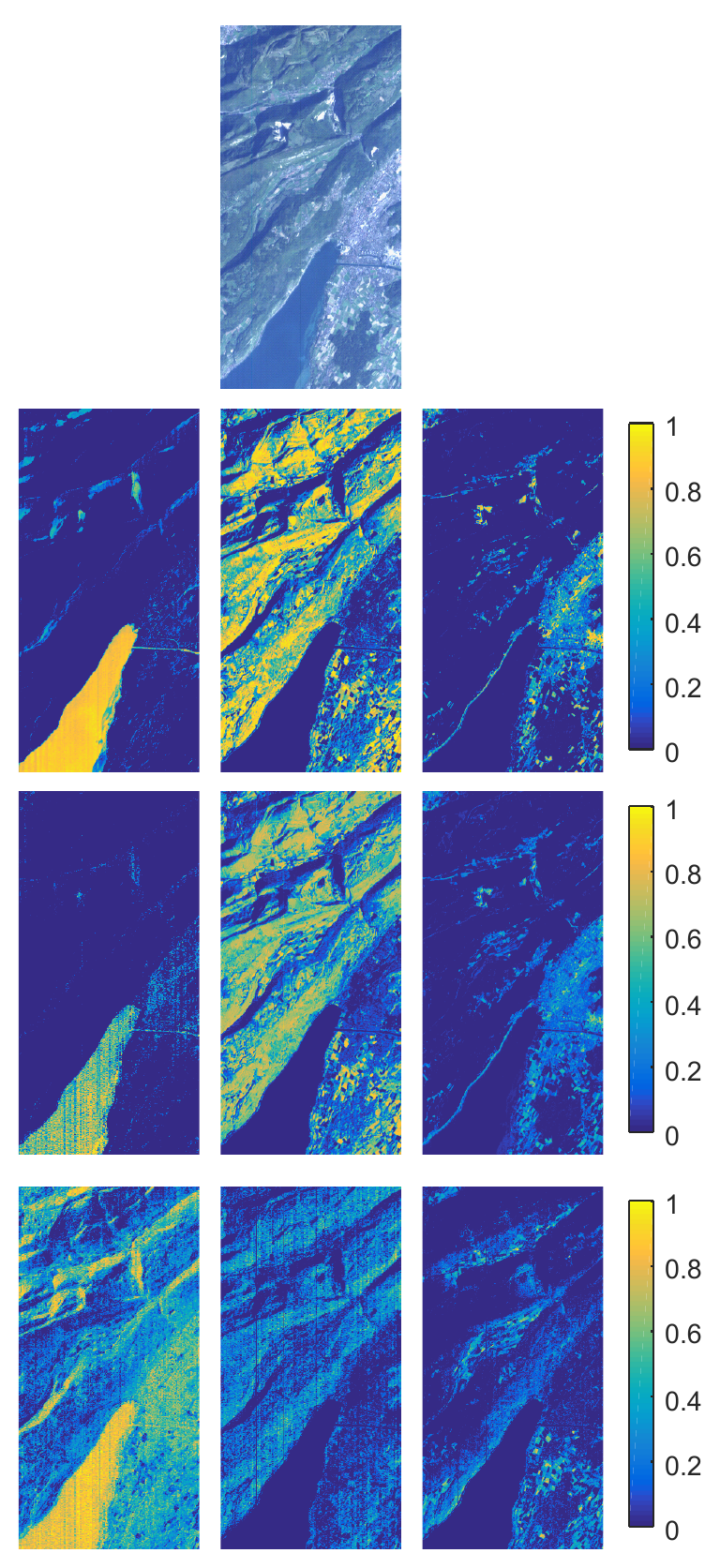}
\caption{Adundances reconstruction from a satellite image. From top to bottom:
input, our prediction, ground truth, tentative denoising over ground truth.
Note how error free and well outlined are our abundances images~\wrt~the ground truth.}
\label{fig:Abund_Hyperion}
\end{center}
\end{figure}

\subsection{Denoising}
An interesting property of our learned upsampling is that it can be
used as a denoising method: downsampling the original images (as we do
in our experiments) removes noise, but upsampling does not re-insert
it. Indeed it is known that deep neural networks achieve
state-of-the-art results in image denoising~\cite{xie2012image}.
See the prediction in Fig.~\ref{fig:noiseNUS}, note how the marked line artifacts in the ground truth get removed by our method.
On the satellite data, which is in general much noisier, this effect
gets very prominent. In most of the cases the predicted images for
Hyperion are cleaner and more useful than the original ``ground
truth'' one.  In Fig.~\ref{fig:hyperion1} the difference images is
dominated by the noise, while the ``true'' prediction error appears
minimal. This claim is further supported by the fact that we were able
to extract plausible spectral endmembers from the predicted
hyperspectral images, which we found impossible for the originals.

\subsection{Hyperspectral Unmixing}
We also check our reconstruction on satellite data, by performing hyperspectral unmixing \cite{bioucas2012hyperspectral} a process that separates material information (also called \emph{endmembers}) and their location in the image (also called \emph{abundances}).
We take an of the shelf endmember extraction algorithm (VCA, \cite{li2008minimum}) to identify dominant spectral signatures in the images. Then, we perform a Fully Constrained Least Squares (FCLS) adjustment to extract the abundance maps, according to the Linear Mixing Model (LMM) \cite{keshava2002spectral}. The abundance maps show the presence of each endmember in each pixel and are constrained to be non-negative and sum to one.
We select a subset of one image and extract 15 endmembers and their corresponding abundances, for three different cases of hyperspectral image: Our prediction, ground truth and ground truth denoised, by projecting the data points onto the first 15 principal components of the image (PCA projection), see Fig~\ref{fig:Abund_Hyperion}. This kind of denoising method is suited for white noise as long as its variance is lower that that of the signal. Unfortunately, this is not enough to remove the noise from the abundance estimation, because the noise in this problem is not white, and so strong that apparently 15 principal components are insufficient to cover the underyling (obviously non-linear) subspace.
As can be seen in Fig.~\ref{fig:Abund_Hyperion} the ground truth itself cannot be used for hyperspectral unmixing as it is noisy.
On the other hand, our method, only using 9 dimensions, is denoising the image as can be seen by the sharp abundance images, which clearly depict water, vegetation and urban areas, second row.









%% file: tableICVL-CAVE.tex
\begin{table*}[t]
  \centering
  \caption{Comparison of our method with Arad~\etal~\cite{arad2016sparse} on ICVL and CAVE dataset.}
\label{table:ICVL}
  \begin{tabular}{l|cc|cc}
    \hline
     & \multicolumn{2}{c|}{ICVL}& \multicolumn{2}{c}{CAVE}\\

   & Ours   & Arad \etal \cite{arad2016sparse} & Ours  & Arad \etal \cite{arad2016sparse} \\
       \hline
RMSE     & \textbf{1.980} & 2.633  & \textbf{4.76} & 5.4 \\
RMSERel & \textbf{0.0587} & 0.0756 & 0.2804          & --  \\
SAM      & 2.04          & --     & 12.10          & --  \\
    \hline
  \end{tabular}
\end{table*}

%% file: tableNUSrad.tex
\begin{table}[t]
\centering
\caption{Error evaluated on 8-bit images over the radiance \wrt~\cite{nguyen2014training} on NUS dataset}
\label{table:NUSradiance}
\begin{tabular}{l|ccc}
\hline
 & RMSE & RMSERel  & SAM\\
\hline
Nguyen \etal \cite{nguyen2014training} & 8.99          & 0.324          & \textbf{9.23}           \\
Ours        & \textbf{5.27} & \textbf{0.234} & 10.11 \\
\hline
\end{tabular}
\end{table}

%% file: tableNUSrefl.tex
\begin{table}[t]
\centering
\caption{Error evaluation on the reflectance using the same procedure as in~\cite{nguyen2014training} on NUS dataset}
\label{table:NUSreflectance}
\begin{tabular}{l|ccc}
\hline
 & RMSE & RMSERel  & SAM\\
\hline
Nguyen \etal \cite{nguyen2014training} & 0.0451 & 0.3070 & \textbf{10.37} \\
Ours        & \textbf{0.0390} & \textbf{0.2406} & 11.94 \\
\hline
\end{tabular}
\end{table}

%% file: tableHyperion.tex
\begin{table}[t]
\centering
\caption{Quantitative evaluation of our method on satellite images on different
dates of the same scene.}
\label{table:Hyperion}
\begin{tabular}{r|ccc}
\hline
                    & RMSE & RMSERel & SAM  \\
\hline
21st March 2014     & 0.54 & 0.090 & 2.90 \\
11th April 2014     & 0.63 & 0.075 & 2.50 \\
24th April 2015     & 0.75 & 0.085 & 2.84 \\
7th May 2015        & 0.95 & 0.114 & 3.69 \\
28th September 2015 & 0.62 & 0.103 & 3.11 \\
8th May 2016        & 1.28 & 0.089 & 3.17 \\
18th July 2016      & 1.06 & 0.149 & 5.04 \\
14th August 2016    & 2.55 & 0.139 & 5.04 \\
\hline
\end{tabular}
\end{table}

%% file: conclusion.tex
\section{Conclusions}
We show that it is possible to do super resolution for image not only in the
spatial domain but also in the spectral domain.
Our method builds on a recent high-performance convolutional neural network, which was originally designed for semantic segmentation.
Contrary to other work on spectral super-resolution, we train and predict directly the end-to-end relation
between an RGB image and its corresponding hyper-spectral
image, without using any additional input, such as the spectral response function.
We show the performance of our work on multiple indoor, outdoor and satellite
datasets, where we compare favorably to other, less generic methods.
We believe that our work may be useful for a number of applications that would benefit from higher spectral resolution, but where the recording conditions or the cost do not allow for routine use of hyper-spectral cameras.